\newif\ifTR
\TRtrue  

\ifTR
\documentclass[onecolumn,12pt]{IEEEtran}
\IEEEoverridecommandlockouts

\else
\documentclass[twocolumn,10pt]{../../IEEEtran}
\fi

\usepackage{url,amsmath,amsfonts,amssymb,mathrsfs,epsfig,bm}
\usepackage{graphicx}

\newtheorem{theorem}{Theorem}[section]

\newcommand{\qed}{\raisebox{.65ex}{\fbox{\rule{0mm}{0mm}}} ~\\}

\newcommand{\hatc}{\hat{c}}

\newcommand{\be}{\begin{eqnarray}}
\newcommand{\ee}{\end{eqnarray}}

\newcommand{\beqa}{\begin{eqnarray*}}
\newcommand{\eeqa}{\end{eqnarray*}}

\newcommand{\cf}{{\em cf., }}
\newcommand{\eg}{{\em e.g., }}
\newcommand{\ie}{{\em i.e., }}

\renewcommand{\th}[1]{{#1}\mbox{th}}

\newcommand{\Tcal}{{\mathcal{T}}}

\def\Reals{{\mathbb{R}}}

\def\B{{\sf B}}

\begin{document}
\title{
Notes on Margin Training and Margin p-Values for Deep Neural Network Classifiers
}

\author{George Kesidis, David~J. Miller, 
and Zhen Xiang\thanks{The authors are with the School of EECS,
Pennsylvania State University,  University Park, PA, 16803, USA.
This research is supported by AFOSR DDDAS grant and Cisco URP gift.
Email: \{gik2,djm25,zux49\}@psu.edu}
}

\maketitle
\begin{abstract}
We provide a new local class-purity theorem for Lipschitz continuous
DNN classifiers. 
In addition, we discuss how to achieve classification 
margin for training samples. 
Finally, we describe how to compute margin p-values for test samples. 
\end{abstract}

\section{Introduction}

Robust DNNs have been proposed to defeat bounded-perturbation
test-time evasion attacks - i.e., small perturbations added
to nominal test samples so that their class decision changes.
One family of approaches
controls Lipschitz-continuity parameter
and targets training-set classification margin.
Estimation and engineering
of the Lipschitz parameter for a DNN is discussed in, \eg 
\cite{Szegedy_seminal,BFT17,Parseval17,Tsuzuku18,Weng18,Gouk18,Pappas19}.
How to 
engineer class purity (class decision consistency) in an 
convex neighborhood (open ball)
of a certain size about every training samples 
is addressed in \cite{Tsuzuku18,Provable}.
In the following, we give an alternative local class
purity result. Also, we 
show how to achieve classification margin on 
training samples
by choice of a simple ``dual" training objective,
\cf (\ref{cel-dual0}) and (\ref{cel-dual1}).
We numerically show how margin-based
training 
can result in reduced accuracy
(by overfitting the training set).
Finally, we define a p-value associated with classification margin.

\section{Margin in DNN classifiers}

Consider the DNN $f:\Reals^n\rightarrow (\Reals^+)^C$  
where $C$ is the number of classes.
Further suppose that for an input pattern $x\in\Reals^{n}$ 
to the DNN,
the class decision is
\beqa
\hatc(x) = \mbox{arg}\max_i f_i(x),
\eeqa
where $f_i$ is the $\th{i}$ component of the $C$-vector $f$.
That is, we have defined a class-discriminant output layer of the DNN.
Here assume that a
class for $x$ is chosen arbitrarily among those that tie for the maximum. 
In the following, we assume that the functions $f_i$ are rectified:
\be\label{rectified}
\forall i,x, ~f_i(x) & \geq & 0.
\ee

Define the {\bf margin} of $x$ as 
\be\label{margin-def}
\mu_f(x)~:=~f_{\hatc(x)}(x)-\max_{i\not=\hatc(x)} f_i(x) & \geq & 0.
\ee

Now suppose the $\ell_{\infty}/\ell_2$
 Lipschitz continuity parameter $L_\infty$ for $f$, \ie
the smallest $L_\infty>0$ satisfying
\be\label{Lip-def}
\forall x,y, ~~|f(x)-f(y)|_\infty & \leq & L_\infty|x-y|_2
\ee
is estimated.  Note that we have used two different norms
in this definition.

Now consider  samples in a open $\ell_2$  
ball centered at $x$, \ie
$$y\in \B_2(x,\varepsilon):=\{z\in\Reals^n~:~|x-z|_2
< \varepsilon\}$$
for $\varepsilon>0$.

The following is a locally consistent (robust) classification result
is an example of {\em Lipschitz margin} \cite{Tsuzuku18}.

~\\

\begin{theorem}\label{thm:local-purity}
If $f$ is $\ell_\infty/\ell_2$ Lipschitz continuous with parameter
$L_\infty >0$ and
$\mu_f(x)>0$ then
$$ \B_2\left( x,\frac{\mu_f(x)}{2L_\infty}\right)$$
is class pure.
\end{theorem}

~\\
\noindent {\bf Proof:}
For any  $y\in \B_2(x,\frac{1}{2}\mu_f(x)/L_\infty)$,  we have
\beqa
\frac{1}{2}\mu_f(x) & >&  L_\infty |x-y|_2\\
& \geq & |f(x)-f(y)|_\infty\\
 & := & \max_i |f_i(x)-f_i(y)|_\infty\\
& \geq & \max_i |f_i(x)|_\infty-|f_i(y)|_\infty 
~~~\mbox{(triangle inequality)}\\
& = & \max_i f_i(x)-f_i(y) ~~~\mbox{(since $f_i\geq 0$)}\\
& \geq & f_{\hatc(x)}(x)-f_{\hatc(x)}(y)
\eeqa
So, 
\be\label{y-bound1}
f_{\hatc(x)}(y) & > & f_{\hatc(x)}(x) -\frac{1}{2}\mu_f(x).
\ee

If we instead write $|f_i(y)|_\infty-|f_i(x)|_\infty$ 
in the triangle inequality above
and then replace $\hatc(x)$ by any $i\not=\hatc(x)$, we get that
\be\label{y-bound2}
\forall i\not=\hatc(x),~~ f_{i}(y) & < & f_{i}(x) +\frac{1}{2}\mu_f(x).
\ee 
So, by  (\ref{y-bound1}) and (\ref{y-bound2}),
\beqa
\forall i\not=\hatc(x),~~
f_{i}(y) & < & f_{i}(x) +\frac{1}{2}\mu_f(x) \\
 & \leq & f_{\hatc(x)}(x) -\frac{1}{2}\mu_f(x) ~~~\mbox{(by (\ref{margin-def}))}\\
 & < & f_{\hatc(x)}(y)
\eeqa
\qed

Theorem \ref{thm:local-purity} is similar to Proposition 4.1 of 
\cite{Tsuzuku18}. Let the 2-norm 
Lipschitz parameter of $f$ be $L_2$, \ie using the 2-norm on
both sides of (\ref{Lip-def}).  
Since  $|z|_\infty \leq |z|_2$ for all $z$, 
$L_2\geq L_\infty.$
Without assuming $f$ is rectified as (\ref{rectified}),
\cite{Tsuzuku18} shows that 
$y$ is assigned the same class as $x$ if
$\mu_f(x)  >   \sqrt{2}L_2|x-y|_2$;
thus, 
$\B_2(x,\mu_f(x)/(\sqrt{2}L_2))$
is class pure. 
Note that  $\sqrt{2}L_2$ 
(Prop. 4.1 of \cite{Tsuzuku18}) may or may not be
larger than $2L_\infty$ (Theorem \ref{thm:local-purity}).
On the other hand, if the right-hand-side of
(\ref{Lip-def}) is changed to the $\ell_\infty$ norm, then using
$|z|_2 \leq n|z|_\infty$ for all $z$,
and arguing as for Theorem \ref{thm:local-purity}) leads
to a weaker result than Prop. 4.1 of \cite{Tsuzuku18}
(especially when $n\gg 1$).

\section{Margin training}\label{sec:lmt}

Robust training is surveyed in \cite{shiqi}.
Lipschitz margin training to achieve a class-pure 
convex neighborhood (open ball) of prescribed size
about every training sample is discussed in \cite{Tsuzuku18},
combining 
margin training (\ref{margin-def}) and
Lipschitz continuity parameter control.
(Also see e.g. \cite{Parseval17} for Lipschitz parameter
control and the approach for  bounding margin gradient 
of \cite{Certified}.)
\cite{Provable} relaxes the constraints of ReLU based
classifiers toward this same objective (assuming ReLU neurons
with bounded outputs).
For a given classifier, the  approach of \cite{Provable}
can also check class purity of a prescribed-size
convex neighborhood of {\em test} samples; using this
method to detect small-perturbation test-time evasion attacks
may have  a significant false-positive rate.
Generally, these methods cannot certify a test sample
is not test-time evasive if the associated perturbation is larger
than the prescribed neighborhood size, and
they may be associated with reduction in classification
accuracy  \cite{Tsuzuku18,Certified}.

We focus herein on just achieving a prescribed 
margin for training samples (\ref{margin-def}).

Let $\theta$ represent the DNN parameters.
Let $\Tcal$ represent the training dataset and let
$c(x)$ for any $x\in\Tcal$ be the {\em ground truth}
class of $x$. The following is easily generalized
to sample-dependent margins ($\mu(x)>0$).

\cite{Tsuzuku18} suggests  to
add the margin  ``to all elements in logits except for the index 
corresponding to" $c(x)$. For example, train the DNN by finding:
\be
\lefteqn{
\min_\theta -\sum_{x\in\Tcal}\log\left(\frac{f_{c(x)}(x)}{\sum_{i\not=c(x)}(f_i(x)+\mu)}\right)} & &  \nonumber \\
& = & 
\min_\theta
-\sum_{x\in\Tcal}\log\left(\frac{f_{c(x)}(x)}{(C-1)\mu+\sum_{i\not=c(x)} f_i(x)}\right) 
\label{tsuzuku-obj}
\ee
For a softmax example,
one could train the DNN using the modified
cross-entropy loss\footnote{Obviously,
exponentiation is unnecessary
when, $\forall x,i$,  $f_i(x)\geq 0$,
\ie the DNN outputs are rectified.}:
\be
\min_\theta -\sum_{x\in\Tcal}\log\left(\frac{\mbox{e}^{f_{c(x)}(x)}}{
\mbox{e}^{f_{c(x)}(x)}+\sum_{i\not=c(x)}\mbox{e}^{f_i(x)+\mu}}\right) 
\label{cel-obj}
\ee
These DNN objectives do not guarantee the margins for all training
samples  will be met.

Alternatively, one can perform
(dual) optimization
of the weighted margin constraints, {\em e.g.},
\be
\min_\theta 
\sum_{x\in\Tcal}
\lambda_x \left(\frac{\max_{i\not=c(x)} f_i(x)+\mu -f_{c(x)}(x)}{(C-1)\mu+\sum_j f_j(x)}\right),
\label{cel-dual0}
\ee
or just
\be
\min_\theta 
\sum_{x\in\Tcal}
\lambda_x \left(\max_{i\not=c(x)} f_i(x)+\mu -f_{c(x)}(x)\right),
\label{cel-dual1}
\ee
where the DNN mappings $f_i$ obviously depend on the DNN parameters
$\theta$, and the weights $\lambda_x\geq 0$ $\forall x\in\Tcal$.
For hyperparameter $\delta>1$, training can proceed simply as:
\begin{itemize}
\item[0] Select initially equal $\lambda_x >0$, say $\lambda_x=1$ $\forall x\in\Tcal$.
\item[1] Optimize over $\theta$ (train the DNN).
\item[2] If all margin constraints are satisfied then stop.
\item[3] For all $x\in \Tcal$: if margin constraint $x$ is
not satisfied then 
$\lambda_x \rightarrow\delta \lambda_x$.
\item[4] Go to step 1.
\end{itemize}
Again, the parameters of the previous 
DNN could initialize the training of the next,
and an initial DNN can be trained instead by 
using a logit or cross-entropy loss
objective, as above.
There are many other variations including also decreasing $\lambda_x$ when
the $x$-constraint is satisfied, or additively (rather than 
exponentially) increasing $\lambda_x$ when they are not, and changing
$\lambda_x$ 
in a way that depends on the degree of the corresponding
margin violation.

Given a thus margin trained classifier, one could estimate
its Lipschitz continuity parameter, \eg \cite{Weng18,Gouk18,Pappas19}, 
and apply  Theorem \ref{thm:local-purity}
or Proposition 4.1 of \cite{Tsuzuku18}  to
determine a region of class purity around each training
sample.

\section{Some Numerical Results for Classification Margin}\label{sec:numer1}

\begin{figure}
\centering
\ifTR
\includegraphics[width=5in]{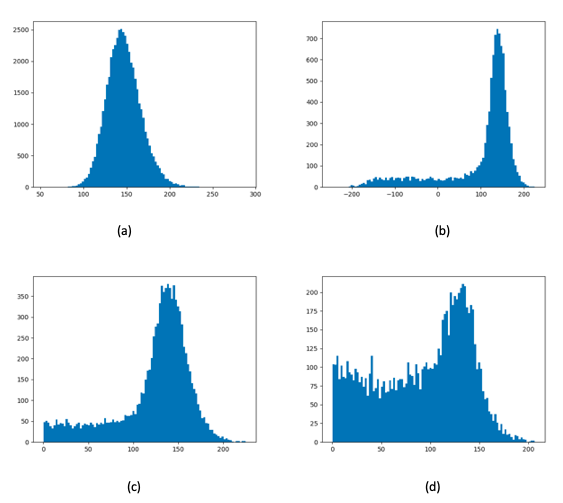}
\else
\fi
\caption{After training using (\ref{cel-dual1}) with margin $\mu=50$,
resulting histogram of margins of: (a) training samples
(b) labelled samples held-out from training dataset;
(c) test dataset (labels unknown, so decisions by the classifier
itself are used to determine margin here); and
(d) FGSM samples created by the test dataset (c).
Note that the sample values in cases (b) and (c) are the same.
}\label{fig:margin50}
\end{figure}

\begin{figure}
\centering
\ifTR
\includegraphics[width=5in]{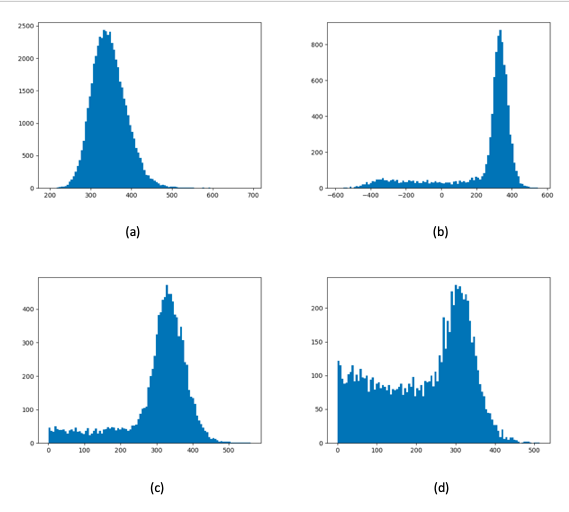}
\else
\fi
\caption{After training using (\ref{cel-dual1}) with margin $\mu=150$,
resulting histogram of margins of: (a) training samples
(b) labelled samples held-out from training dataset;
(c) test dataset (labels unknown, so decisions by the classifier
itself are used to determine margin here); and 
(d) FGSM samples created by the test dataset (c).
Note that the sample values in cases (b) and (c) are the same.
}\label{fig:margin150}
\end{figure}

In this section, we give an example using loss function
(\ref{cel-dual1}).
Training was performed on CIFAR-10 (50000 training samples and 10000 test/held-out samples) using the ResNet-18 DNN 
(ReLU activations are not used after the fully connected layer).
The training was performed for 200 epochs  using a batch size 32 and
learning rate $10^{-4}$.
The results for margins $\mu=50$ and $\mu=150$
are given in Figures \ref{fig:margin50},\ref{fig:margin150} and
Table \ref{tab:accuracy}.

All training-sample
margins were achieved with one training pass using initial
$\lambda_x=1$ for all $x\in\Tcal$;
see Figures \ref{fig:margin50}(a) and \ref{fig:margin150}(a).
 Figures \ref{fig:margin50}(b) and \ref{fig:margin150}(b) show
the margins of the dataset held out from training, i.e., to compute
the margins, the true class label was used. Here, one can clearly
see that many test samples have margins less than $\mu$ and some are
misclassified (negative margins), \cf Table \ref{tab:accuracy}.
 Figures \ref{fig:margin50}(c) and \ref{fig:margin150}(c) show the
margins based on the class decisions of the classifiers themselves,
as would be the case for unlabelled test samples (so all 
measured margins are not negative).
The held-out set and test set are the same.
Finally, 
 Figures \ref{fig:margin50}(d) and \ref{fig:margin150}(d) show the margins
of FGSM \cite{Goodfellow} adversarial samples with parameter/strength
$\varepsilon=0.1$ created using a surrogate  ResNet-18 DNN of the
same structure trained using standard cross-entropy loss
(all such samples were used, including those based on the
$13.3\%$ of test samples that were misclassified).

In Table \ref{tab:accuracy}, we show the accuracy of the
classifiers, including a baseline classifier trained using
the same dataset and ResNet-18  DNN structure but with 
standard cross-entropy loss objective.
As Figures \ref{fig:margin50}(d) and \ref{fig:margin150}(d), the
accuracy performance reported here is for  FGSM adversarial 
samples that were crafted assuming the attacker
knows the baseline DNN trained
by cross-entropy loss. These attacks are transferred to the margin-trained
classifiers.

\begin{table}
\centering
\begin{tabular}{|c||c|c|c|}
\hline
training   & x-entropy & margin    & margin    \\ 
objective $\rightarrow$ & loss      &  $\mu=50$ & $\mu=150$ \\ \hline\hline
clean test-set     &  86.70\% & 85.49\% &  85.37\% \\ \hline
FGSM attacks    &  6.017\% & 10.08\% &  10.08\% \\ \hline
\end{tabular}
\vspace{0.125in}
\caption{Test-time accuracy. Note that the FGSM attacks with parameter/strength
$0.1$ were created using the DNN trained with cross-entropy loss, and transferred
to the margin trained DNNs. The FGSM attacks were based on all test samples
including the 13.3\% that were misclassified by the DNN trained
by cross-entropy loss.}\label{tab:accuracy}
\end{table}

\section{Low-margin atypicality of test samples}\label{sec:margin-atypical}

Given an arbitrary DNN 
$f:\Reals^n\rightarrow (\Reals^+)^C$ ,
let $\Tcal_\kappa$ be the (clean) training samples of class 
$\kappa\in\{1,2,...,C\}$,
\ie $\forall x\in\Tcal_\kappa$, $\hatc(x)=c(x)=\kappa$.
Recall (\ref{margin-def}) and suppose a Gaussian Mixture Model
(GMM) is learned using the log-margins of the training dataset
$$\{\log \mu_f(x)~:~x\in\Tcal_\kappa\}$$
by EM  \cite{Dempster1977} using BIC model order control \cite{Schwarz1978}
as, \eg \cite{MLSP18-PCAD}.
(Instead of margin (\ref{margin-def}), one could use
an estimate the radius of the largest $\ell_2$ ball of
class purity about each training and test sample, e.g.,
directly \cite{Provable} or via estimated Lipschitz constant
as discussed above.)
Let the resulting GMM
parameters be $\{w_i,m_i,\sigma_i\}_{k=1}^{I_\kappa}$, where
$I_\kappa\leq |\Tcal_\kappa|$ is the number of components,
the $w_i\geq 0$ are their weights ($\sum_{i=1}^{I_\kappa} w_i = 1$),
the $m_i$ are their means, and 
the $\sigma_i>0$ are their standard deviations.
So, we can simply
compute the {\bf margin p-value} of any {\bf test sample} $x$,
\beqa
\pi_f(x) & = & \sum_{i=1}^{I_\kappa}  w_i 
\left(1-F\left(\frac{|\log(\mu_f (x)) -m_i|}{\sigma_i}\right)\right)
\eeqa
where $F$ is the standard normal c.d.f.
That is,
$\pi_f(x)$  is the probability that a randomly chosen sample
from the same distribution as that of the training samples
has smaller margin 
than the test sample $x$. So, one can can compare
 $\pi_f(x)$ to a threshold to detect
 whether
a test sample $x$ has abnormally small classification margin.
The example of margin-trained DNN of 
Figures \ref{fig:margin50}(a) and \ref{fig:margin150}(a)
has a single component  for the entire
training set $\Tcal = \bigcup_{\kappa=1}^C \Tcal_\kappa$.
In an unsupervised fashion, the threshold criterion could be
a bound on false positives based on the training set.
Alternatively, the threshold could  be set by using
a clean set of labelled samples that 
were held out from (not used for) training and
consider both false-positive and false-negative performance.

\bibliographystyle{plain}
\bibliography{../../latex/computing,../../latex/ddos,../../latex/ids,../../latex/adversarial,../../latex/kesidis-prior,../../latex/MyCollection,../../latex/refs,../../latex/ref,../../latex/botsalting,../../latex/gans,../../latex/transductive}

\begin{thebibliography}{10}

\bibitem{BFT17}
P.~Bartlett, D.~Foster, and M.~Telgarsky.
\newblock {Spectrally-normalized Margin Bounds for Neural Networks}.
\newblock In {\em Proc NIPS}, 2017.

\bibitem{Parseval17}
M.~Cisse, P.~Bojanowski, E.~Grave, Y.~Dauphin, and N.~Usunierr.
\newblock {Parseval Networks: Improving Robustness to Adversarial Examples}.
\newblock In {\em Proc. ICML}, 2017.

\bibitem{Dempster1977}
Arthur~P. Dempster, Nan~M. Laird, and Donald~B. Rubin.
\newblock {Maximum likelihood from incomplete data via the EM algorithm}.
\newblock {\em Journal of the Royal Statistical Society.}, 39(1):1--38, 1977.

\bibitem{Pappas19}
M.~Fazlyab, A.~Robey, H.~Hassani, M.~Morari, and G.J. Pappas.
\newblock {Efficient and Accurate Estimation of Lipschitz Constants for Deep
  Neural Networks}.
\newblock https://arxiv.org/pdf/1906.04893.pdf, 2019.

\bibitem{Goodfellow}
I.~Goodfellow, J.~Shlens, and C.~Szegedy.
\newblock Explaining and harnessing adversarial examples.
\newblock In {\em {Proc. ICLR}}, 2015.

\bibitem{Gouk18}
H.~Gouk, E.~Frankeib, and B.~Pfahringer.
\newblock {Regularisation of Neural Networks by Enforcing Lipschitz
  Continuity}.
\newblock https://arxiv.org/pdf/1804.04368.pdf, Sept. 2018.

\bibitem{Provable}
J.~Kolter and E.~Wong.
\newblock {Provable defenses against adversarial examples via the convex outer
  adversarial polytope}.
\newblock In {\em Proc. ICML}, 2018.

\bibitem{MLSP18-PCAD}
D.J. Miller, Z.~Qiu, and G.~Kesidis.
\newblock {Parsimonious Cluster-based Anomaly Detection (PCAD)}.
\newblock In {\em Proc. IEEE MLSP}, Aalborg, Denmark, Sept. 2018.

\bibitem{Certified}
A.~Raghunathan, J.~Steinhardt, and P.~Liang.
\newblock {Certified Defenses against Adversarial Examples}.
\newblock In {\em Proc. ICLR}, 2018.

\bibitem{Schwarz1978}
Gideon Schwarz.
\newblock {Estimating the dimension of a model}.
\newblock {\em Annals of Statistics}, 6(2):461--464, 1978.

\bibitem{Szegedy_seminal}
C.~Szegedy, W.~Zaremba, I~Sutskever, J.~Bruna, D.~Erhan, I.~Goodfellow, and
  R.~Fergus.
\newblock Intriguing properties of neural networks.
\newblock In {\em {Proc. ICLR}}, 2014.

\bibitem{Tsuzuku18}
Y.~Tsuzuku, I.~Sato, and M.~Sugiyama.
\newblock {Lipschitz-margin Training: Scalable Certification of Perturbation
  Invariance for Deep Neural Networks}.
\newblock In {\em Proc NIPS}, 2018.

\bibitem{shiqi}
S.~Wang, Y.~Chen, A.~Abdou, and S.~Jana.
\newblock {MixTrain: Scalable Training of Verifiably Robust Neural Networks}.
\newblock https://arxiv.org/abs/1811.02625, Nov. 2018.

\bibitem{Weng18}
T.-W. Weng, H.~Zhang, H.~Chen, Z.~Song, C.-J. Hsieh, D.~Boning, I.S. Dhillon,
  and L.~Daniel.
\newblock {Towards Fast Computation of Certified Robustness for ReLU Networks}.
\newblock In {\em {Proc. ICML}}, 2018.

\end{thebibliography}
\end{document}